\documentclass[conference]{IEEEtran}
\IEEEoverridecommandlockouts
\usepackage{cite}
\usepackage{amsmath,amssymb,amsfonts}
\usepackage{algorithmic}
\usepackage{graphicx}
\usepackage{textcomp}
\usepackage{xcolor}
\usepackage{hyperref}
\usepackage{booktabs}
\usepackage{array}

\hypersetup{ colorlinks, citecolor=black, linkcolor=red, urlcolor=blue}

\def\BibTeX{{\rm B\kern-.05em{\sc i\kern-.025em b}\kern-.08em
    T\kern-.1667em\lower.7ex\hbox{E}\kern-.125emX}}
\begin{document}

\newcolumntype{C}[1]{>{\centering\arraybackslash}m{#1}}

\title{AI-based Wildfire Prevention, Detection and Suppression System
}

\author{Prisha Shroff%
\thanks{Prisha Shroff is with Hamilton High School, Chandler, AZ.}
\thanks{email: \url{prishashroff@gmail.com} }%
}

\maketitle

\begin{abstract}
Wildfires pose a serious threat to the world’s environment. The global wildfire season length has increased by 19\% and severe wildfires have besieged nations around the world. Every year, forests are burned by these wildfires, causing vast amounts of carbon dioxide to be released into the atmosphere, contributing to climate change. 
Therefore, there is a need for a system which prevents, detects, and suppresses wildfires. 
The AI-based Wildfire Prevention, Detection and Suppression System (WPDSS) is a novel, fully-automated, end-to-end, AI-based solution to effectively predict hotspots (areas where wildfires can start) and detect wildfires, deploy drones to spray fire retardant, preventing and suppressing wildfires. WPDSS consists of four steps.
1) Pre-processing: WPDSS loads real-time satellite data from NASA and meteorological data from NOAA of vegetation, temperature, precipitation, wind, soil moisture, and land cover for prevention. For detection, it loads the real-time data of Land Cover, Humidity, Temperature, Vegetation, Burned Area Index, Ozone, and CO2. It uses the process of masking to eliminate not-hotspots and not-wildfires such as water bodies, and rainfall. 
2) Learning: The AI model consists of a random forest classifier, a type of supervised machine learning, which uses a series of decision trees to make an accurate decision. It is trained using a labeled dataset of hotspots/wildfires and not-hotspots/not-wildfires. 
3) Identification of hotspots and wildfires: WPDSS runs the real-time data through the model to automatically identify hotspots and wildfires.
4) Drone deployment: The drone flies to the identified hotspot/wildfire location.
WPDSS attained a 98.6\% accuracy in identifying hotspots and a 98.7\% accuracy in detecting wildfires. 
WPDSS will reduce the impacts of climate change, protect ecosystems and biodiversity, avert huge economic losses, and save human lives. The power of WPDSS developed can be applied to any location globally to prevent and suppress wildfires, reducing climate change.
\end{abstract}

\begin{IEEEkeywords}
machine learning, wildfires, prevention, detection, suppression
\end{IEEEkeywords}
\vspace{-2mm}

\section{Introduction}
\subsection{Problem}
Wildfires pose a serious threat to the world’s environment. Forests comprise more than 31\% of the world’s land surface and contribute to the continuity of ecological balance, and they store large amounts of carbon. When forests burn, they release carbon dioxide into the atmosphere, contributing to climate change. 

In the last few years, severe wildfires have besieged nations around the world, including the United States, Canada, Australia, Greece, Spain, Portugal, Chile, Russia, China, Siberia, South Korea, Israel and Brazil. For example, the 2019–2020 wildfire season in Australia caused the burning of over 46 million acres of forest while destroying over 10,000 structures, resulting in over \$100 billion in damages. 

According to the World Wildlife Fun, the length of the global fire season has increased on average by 19\% . As stated by the National Interagency Fire Center, there were around 58,250 wildfires and more than 10.3 million acres burned in the US in 2020. The costs for 2020 wildfires in the Western US alone totaled between \$130-\$150 billion according to AccuWeather. The lasting effects of wildfires include ecosystem and biodiversity loss, forest and soil degradation, air pollution, economic losses, destruction of watersheds, and negative impacts to human health. \cite{A2013TheW}.

As the world continues to get warmer and drier, experts believe that the intensity of wildfires will increase in areas like Australia, California, and the Mediterranean. Recently, fires above the Arctic Circle have burned vast swaths of typically frozen, carbon-rich peat soils, releasing a record 244 megatons of carbon into the atmosphere. This is more than Spain releases from burning fossil fuels in an entire year, according to Copernicus Atmosphere Monitoring Service. 

There is a need for a system that can effectively predict hotspots and detect wildfires, to prevent and suppress these wildfires. 

\subsection{Current Solutions}
The current solutions in the market do not have wildfire prevention and detection capabilities. A couple of the solutions are described as follows:
1. Fire Urgency Estimator in Geosynchronous Orbit. This system detects actual fires using satellite images, and uses drones to track the fire’s progress. The solution doesn’t prevent the wildfire from occurring.
2. Wildfire Detection System InsightFD + Insight Globe. This system consists of a network of robots that are constantly monitoring areas with thermal imaging and cameras. Once it detects a wildfire, it alerts and notifies the fire department. The solution doesn’t prevent wildfires from occurring and requires robots to be installed at each site.
3. BurnMonitor: An Early Wildfire Detection IoT Solution. This system consists of a network of sensors that can detect if a fire is happening nearby. It sends a signal to a gateway which transmits the location to the fire department. The solution doesn’t prevent wildfires from starting but instead detects the fire.
4. San Diego Gas and Electric iPredict.  This system uses an AI based predictive model to increase the accuracy of weather forecasts and predicts powerline caused wildfires. It only prevents these powerline caused wildfires, and does not detect wildfires. 
5. Smart Wildfire Sensor Device. This uses sensors that are installed on the site to predict wildfires. It analyzes the amount of dead fuel, the moisture of the fuel and other datasets to predict the wildfires, but the biggest problem with this solution is that it has to be manually installed in every location, which increases the cost exponentially. It also only uses biomass as a factor for prediction, and does not currently detect wildfires. 
6. Specific tailor-made wildfire prevention algorithms have been created in regions such as Spain \cite{AlonsoBetanzos2003AnIS}, France \cite{Jov2014FireFightAD}, Turkey \cite{Gumusay2009VisualizationOF}, the US \cite{NoonanWright2011DevelopingTU}, and in Europe \cite{Corgnati2008FIREcastS}. However, these systems can predict wildfires, but they are not end-to-end solutions that take action to prevent the wildfires from starting (removing fuel, spraying fire retardant, etc.). They are region-specific (use only that region’s data) and have no wildfire detection capabilities.
After looking at the current solutions developed by the scientific community, it is clear that there is a need for a system that can predict hotspots, detect wildfires and send a drone to the location to prevent and suppress wildfires. Prediction and detection of these wildfires can help prevent and mitigate the disastrous effects.

\section{Design}

\begin{figure*}[tbh!]
\centerline{\includegraphics[width=0.95\textwidth]{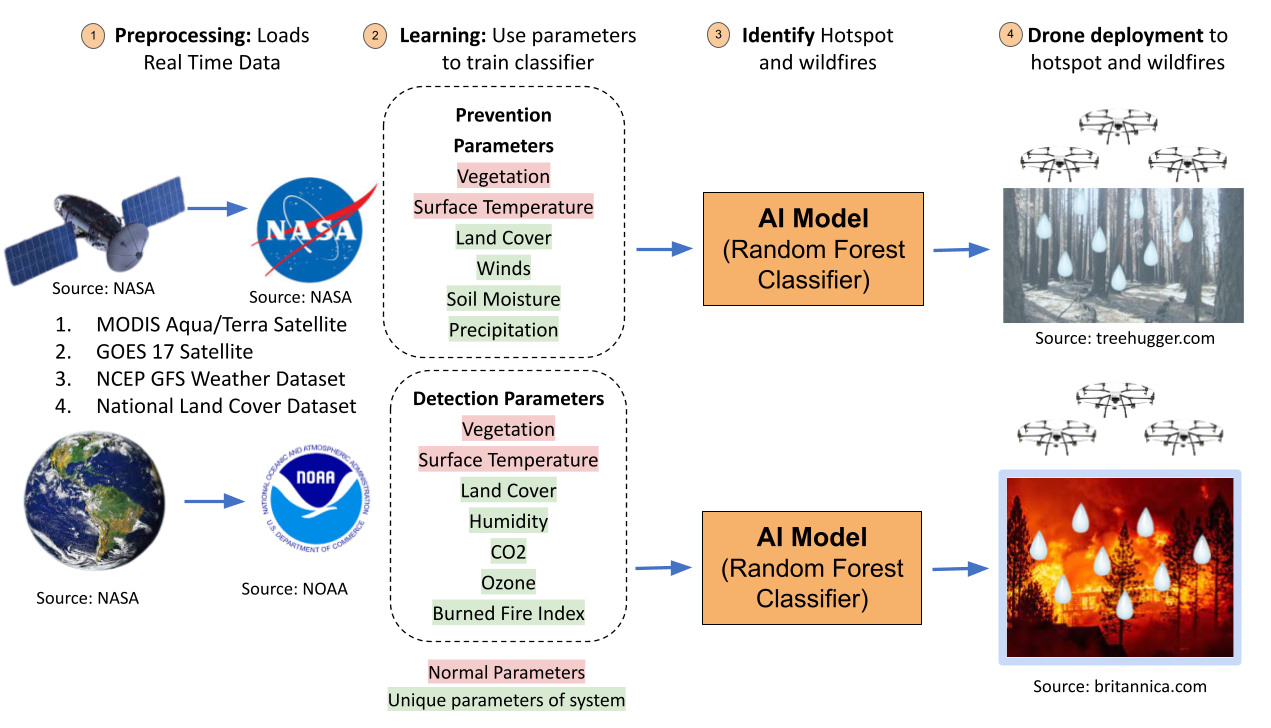}}
\caption{High-level overview of WPDSS.}
\label{fig1}
\vspace{-5mm}
\end{figure*}

\subsection{Design Criteria}
The AI-based Wildfire Prevention, Detection and Suppression System should be able to use real-time satellite data and artificial intelligence to: automatically identify hotspots, automatically identify current wildfires, attain higher than 90\% accuracy in hotspot and wildfire identification and automatically notify a drone in the area to fly to the hotspot or current wildfire.

\subsection{Artificial Intelligence}
The Wildfire Prevention, Detection and Suppression System identifies hotspots and wildfires using artificial intelligence. A few terms regarding artificial intelligence are clarified. 
1. Machine Learning:  Machine learning is an intelligent system with a human-like brain that learns through data. A branch of machine learning is supervised machine learning. Supervised learning involves giving the algorithm training datasets with labels. Once familiarized, it is given real-time data which it labels on its own.
2. Masking: Masking is the process of eliminating areas where wildfires cannot and will not occur. These specific areas are not included in the AI Random Forest Model. Masking is performed on areas with rainfall, ice, oceans, etc. 
3. Underfitting: Underfitting is when the model performs poorly on the training data because the model is unable to capture the relationship between the input examples (X) and the target values (Y).
4. Overfitting: Overfitting is when the model performs well on the training data but does not perform well on the evaluation data. This is because the model is memorizing the data it has seen and focuses on patterns relative to the dataset. When the model is tested on a new dataset, it cannot reapply the patterns it originally formulated. 
5. Evaluation Metrics: Precision shows out of the number of times a model predicted 0 or 1, how often was it correct? Recall shows out of the total actual 0 or 1, how many of them were correctly predicted as 0 or 1? 

\subsection{Parameters}

The Wildfire Prevention, Detection and Suppression System uses a series of model parameters to accurately predict hotspots and identify wildfires. To predict hotspots, WPDSS used data of land cover, wind speed, precipitation, soil moisture, temperature, and vegetation as input to the model. The satellite and meteorological data was loaded from the National Land Cover Dataset (NCLD), NCEP GFS Weather (NOAA), and MODIS Aqua/Terra Satellite (NASA) of parameters. 
1) Land Cover is the type of land surface. The dataset provides data on thematic class (for example, urban, agriculture, and forest), percent impervious surface, and perfect tree canopy cover. WPDSS uses land cover to mask areas with open water, perennial ice/snow, developed land (open space, low, medium and high intensity), barren land (rock, sand, clay), moss, woody wetlands and emergent herbaceous wetlands. 
2) Wind Speed shows the speed of the wind in meters per second. Wind plays a major role in the determination of hotspots and wildfires as strong gusts of wind can spread wildfire flames quickly. 
3) Precipitation rate is the amount of rainfall in an area measured in kg/(m\^2 s). Precipitation is shown to have a link with wildfires, as the lack of precipitation has correlated with increased wildfire risk. \cite{Chen2014TheIO}
4) Soil Moisture is the volumetric soil moisture content around 0.0-0.1 m depth below the land layer level. Soil moisture affects wildfires by controlling biomass growth, fire fuel availability, and determines vegetation moisture content (O, S, 2020).
5) Temperature measures the amount of hotness or coldness. This temperature dataset consists of 2 datasets which are combined for accurate depictions of temperature. 
6) Vegetation is the amount of greenery of an area. In this model, the Normalized Difference Vegetation Index (NDVI) is a dataset that estimates the density of green on an area of land. The dataset formula is [ndvi = (nir - red) / (nir + red)].
Whereas, to detect wildfires, WPDSS uses data of land cover, humidity, temperature, vegetation, ozone, CO2 and burned area as input to the model.  The satellite and meteorological data was loaded from the GOES 17 Satellite (NASA and NOAA), National Land Cover Dataset (NCLD), NCEP GFS Weather (NOAA), and MODIS Aqua/Terra Satellite (NASA) of parameters. 
The model parameters of land cover, vegetation, and temperature are the same as the prevention model. The other 4 parameters are described as follows:
1) Humidity is the amount of water vapor present in the air. The dataset consists of specific humidity at the level of 2 meters above ground. 
2) Burned area shows the burnt land. This consists of 2 datasets Near Infrared and short-wave infrared combined using the formula bai = (nir - swir) / (nir + swir). 
3) Ozone is a product of wildfires and can be used to detect wildfires. 
4) CO2, carbon dioxide, is emitted through wildfires. When forests burn, all the CO2 stored in the trees gets released into the atmosphere. 

\subsection{Determination of Model}
To determine the type of method used for the prediction of hotspots and detection of wildfires, three different models were tested, and the one with the highest accuracy was used in the classifier. 
One of the first methods tested used OpenCV and image processing to identify hotspots and wildfires. When using open CV and image processing, an accuracy of 33.3\% was obtained. 
Logistic regression is a statistical analysis method used for binary classification based on a dataset. A logistic regression model classifies the category by analyzing the relationship between the model parameters.  Based on the historical data and previous hotspot and wildfire example, it can accurately classify new data. When tested with the logistic regression model, an accuracy of 74.8\% was attained.
A random forest classifier is a binary classification model that uses a series of decision trees to make the most accurate decision. Decision trees are a method used by an algorithm in which it uses a simple set of decisions to classify the input. The random forest model uses multiple decision trees and chooses the answer that has appeared the most times to get the most accurate result.  To get the most accurate results, parameters can be tuned such as n-estimators (the number of decision trees in the classifier) and max-depth (the number of sequential branches on the decision tree). When tested with the random forest classifier, an accuracy of 96.6\% was attained.
After these tests, it was concluded that the Random forest classifier was the best method for identification of hotspots and wildfires.

\subsection{Training of WPDSS}

To create and train WPDSS, the labeled training/testing dataset was created by manually identifying hotspots, not-hotspots, wildfires, and not wildfires using the model parameters. (one-time step for supervised machine learning) Then, the dataset is split into training (80\%) and testing (20\%). The training dataset is supplied as input to the Random Forest Classifier to learn the hotspot and wildfire occurrence patterns. The testing dataset is supplied as input to the random forest classifier to classify areas as hotspots, not-hotspots, wildfires and not wildfires. Then, tune the n-estimators (the number of decision trees in the random forest classifier) and max\_depth (the number of branches in a decision tree) parameters to increase classifier accuracy. This is completed through a series of 3 steps: 1) modify n-estimators and max\_depth from 1-15 and store the test and train accuracy, 2) determine the highest model accuracy from the data collected and then 3) evaluate the classifier using accuracy, confusion matrix, precision and recall.

\vspace{-1mm}
\subsection{WPDSS Methods}

WPDSS is a combined system that consists of 2 separate random forest classifiers. One of these algorithms predicts hotspots and the other detects wildfires. It performs 4 main steps: preprocessing, learning, identification of hotspots and wildfires, and drone deployment.  
First, WPDSS loads the real-time NASA satellite data and NOAA meteorological data of the parameters to enhance accuracy and precision of hotspot and wildfire detection. It uses the process of masking to eliminate not-hotspots such as water bodies and rainfall to limit the dataset size. WPDSS then runs the real-time data (with the masking) through the model to identify hotspots and wildfires. It removes only the hotspots where wildfire retardant is not allowed to be sprayed. It sends the hotspot and wildfire coordinates to the drone. Then, a prototype tello drone is deployed to the hotspots.

\begin{figure}[htbp]
\centerline{\includegraphics[width=\linewidth]{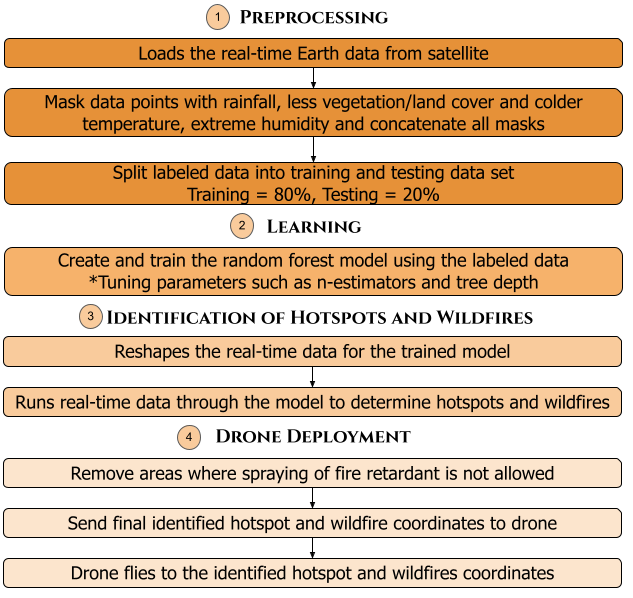}}
\caption{Flow Chart}
\label{fig2}
\vspace{-3mm}
\end{figure}

\subsection{Drones}
The solution improves upon current technology for prevention and suppression because rather than using humans and helicopters, the system will be using drones. Drones are automated and more cost-efficient than firefighters spraying fire retardant in a location. Also, there are helicopters which can spray fire retardant, but require a pilot, cannot spray in the night, costs millions of dollars and cannot fly low.

\section{ANALYSIS}
\subsection{Tuning of Random Forest Classifier Parameters}

\vspace{-1mm}
\begin{figure}[!h]
\centerline{\includegraphics[width=0.95\linewidth]{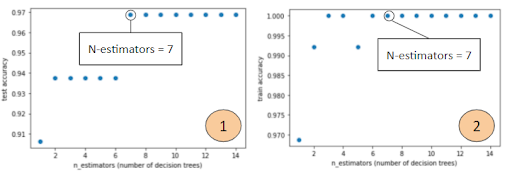}}
\vspace{-2.5mm}
\caption{Tuning of N\_estimators}
\label{fig3}
\vspace{-3mm}
\end{figure}

 There were a few patterns noticed in the data as seen in Fig. \ref{fig3}. When the number of n-estimators < 7, the model is underfitting the data. When the number of n-estimators > 7, the model is overfitting the data. From graphs 1 and 2, it was analyzed that for the highest model accuracy, the number of n-estimators should be 7. 

\begin{figure}[htbp]
\centerline{\includegraphics[width=\linewidth]{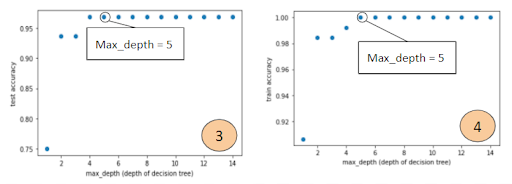}}
\caption{Tuning of max\_depth}
\label{fig4}
\vspace{-2mm}
\end{figure}
 
There were some patterns noticed in these 2 graphs as seen in Fig. \ref{fig4}. When the max\_depth < 5, the model is underfitting the data. When the max\_depth > 5, the model is overfitting the data. From graphs 3 and 4, it was analyzed that for the highest model accuracy, the max\_depth should be 5.

\subsection{Statistical Methods}

In total, 378 labeled coordinates were used for the training and testing of the wildfire prevention model, 189 being hotspots and 189 being not-hotspots. The entire dataset was split into a training  (80\% → 302) and a testing dataset (20\% → 76). Whereas, for wildfire detection, 158 labeled coordinates were used for the training and testing. In this dataset, there were considerably fewer training data points of wildfires, and more that were not wildfires since this mimics the real-life scenario. In real life, the majority of the areas will not be wildfires, and only a small portion will be wildfires. This dataset was split into a training dataset (80\% → 126) and a testing dataset (20\% → 32). 
The statistical methods used to analyze the data were accuracy (training, testing and cross-validation), confusion matrix, precision and recall. These are the standard methods for analyzing a random forest AI classifier.

\subsection{Accuracy}

\begin{figure}[htbp]
\centerline{\includegraphics[width=\linewidth]{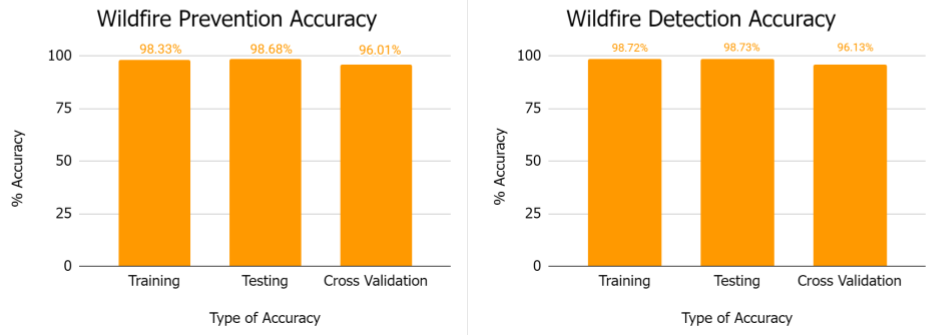}}
\caption{Accuracy}
\label{fig5}
\end{figure}

WPDSS was able to attain 98.6\% testing accuracy in identifying the hotspots for wildfire prevention, when the n-estimators is 7 and max depth is 5. WPDSS was able to attain 98.7\% testing accuracy in detecting wildfires. Cross Validation accuracy for both wildfire prevention and detection is 96\%. The formula for accuracy used is: TP+TN/TP+TN+FP+FN.

Cross validation is a metric used to evaluate machine learning models with a limited training dataset. It has a parameter called k, which is the number of groups that the dataset is split into and k is dependent on the training dataset size. The process of testing the data on unseen data to estimate how the model will perform is called k-fold cross validation. 

\subsection{Confusion Matrix}

\begin{figure}[htbp]
\centerline{\includegraphics[width=\linewidth]{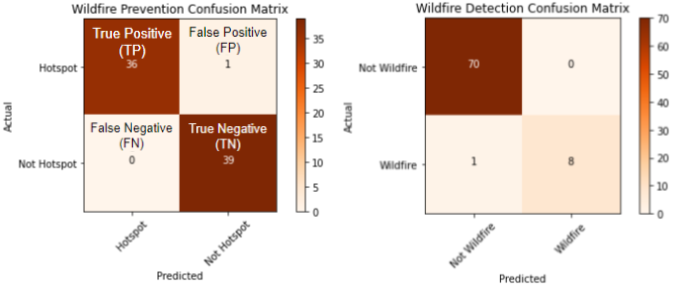}}
\caption{Confusion Matrix}
\label{fig6}
\vspace{-2mm}
\end{figure}

Out of the 76 sample coordinates of the testing dataset for wildfire prevention, 75 were classified correctly. WPDSS predicted the 36 true positives (identified as a hotspot, and it is a hotspot) and 39 true negatives (not identified as a hotspot and is not a hotspot). It predicted 1 false positive (not a hotspot but identified as a hotspot), and 0 false negatives (hotspots being missed). 
Out of the 79 sample coordinates of the testing dataset for wildfire detection, 78 were classified correctly. WPDSS predicted the 70 true positives (identified as a wildfire, and it is a wildfire) and 8 true negatives (not identified as a wildfire and is not a wildfire). It predicted 1 false negative (wildfire being missed), and 0 false positives (not a wildfire but identified as a wildfire).

\subsection{Precision and Recall}

\begin{figure}[htbp]
\centerline{\includegraphics[width=\linewidth]{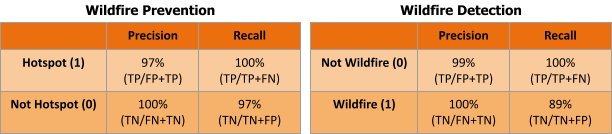}}
\caption{Precision and Recall}
\label{fig7}
\vspace{-2mm}
\end{figure}

WPDSS was 98.5\% precise while identifying a hotspot or not hotspot and was able to recall its previous accurate decisions 98.5\% of the time. WPDSS was 99.5\% precise while identifying a wildfire and was able to recall its previous accurate decisions 94.5\% of the time.

\subsection{Drone Analysis}
Analysis based on use of DJI Agras MG-1 drone to spray fire retardant at a hotspot coordinates

\begin{figure}[htbp]
\centerline{\includegraphics[width=\linewidth]{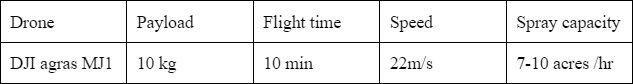}}
\caption{Comparison of Drones}
\label{fig8}
\vspace{-2mm}
\end{figure}

Total distance round trip to spray a hotspot coordinate\\
= 1.0 degree latitude (111 m) x 1.0 degree longitude (111 m) \\
= 12321 m² / 4047 (conversion to acres: 1 acre = 4047 m²)\\
= 3.04 acres \\
Drone’s Spraying Capacity (If drone is carrying 10 kg payload, flight time is reduced to 10 minutes) \\
= 10 acres per hour (1 hour = 60 min)\\
= 1.66 acres per 10 minutes \\
\\
To conclude, number of drones to spray 1 hotspot\\
= 3.04 acres / 1.66 acres \\
= 2 drones for spraying to be completed in 10 minutes OR 1 drone taking 2 round trips to complete spraying in 30 minutes \\
 For Wildfire Prevention and Detection/Suppression,* the above analysis is used for drone deployment. \\
* For detection/suppression rather than using the area in part a, leverage the perimeter to prevent the wildfire from spreading.\\

\section{CONCLUSION}
The AI-based Wildfire Prevention, Detection and Suppression System successfully meets the design criteria. It uses real-time satellite data from NASA and NOAA of soil moisture, precipitation, vegetation, wind speed, land cover, and temperature to identify hotspots using artificial intelligence with a 98.6\% accuracy. It uses real-time satellite data from NASA and NOAA of humidity, CO2, ozone, burned area, vegetation, land cover, and temperature to detect wildfires using artificial intelligence with a 98.7\% accuracy. This system notifies drones to fly to the hotspot coordinates. It was also theoretically proved that the usage of a DJI Agras MG-1 Drone is sufficient enough to spray fire retardant at 1 hotspot coordinate.
The AI-based Wildfire Prevention, Detection and Suppression System can save precious lives, preserve and protect ecosystems and biodiversity and avert huge economic losses (e.g. in the US alone, ~ 10.3 million acres from burning and  ~ $130-$150 billion annually) and help rescue the impacts of climate change,. 
The power of the AI-based Wildfire Prevention, Detection and Suppression System developed can be applied to any location in the world to identify hotspots and detect wildfires, deploy drones to the hotspots and wildfires and therefore prevent wildfires and suppress wildfires.

\section{ FUTURE RESEARCH}
After discussions with Stanford professor Eric Appel, who has built an environmentally friendly fire retardant called fortify which can prevent fire ignition for 6 months, Fortify will be used in WPDSS as the fire retardant. Also, Intel has demonstrated the ability to deploy and control two thousand drones simultaneously. Using the capability of controlling Intel’s drones simultaneously, the wildfires can be prevented and suppressed at a faster speed. Integrating all these efforts and creating a low cost system using these AI based algorithms, wildfire can be prevented and suppressed. 

\bibliographystyle{IEEEtran}
\bibliography{references}

\end{document}